\documentclass[a4paper]{article}

\usepackage{INTERSPEECH2018}

\usepackage[utf8]{inputenc} 
\usepackage[T1]{fontenc}    
\usepackage{hyperref}       
\usepackage{url}            
\usepackage{booktabs}       
\usepackage{amsfonts}       
\usepackage{nicefrac}       
\usepackage{microtype}      
\usepackage{graphicx}
\usepackage{multirow}
\usepackage{color}
\usepackage{amsmath,amssymb,amsthm}
\usepackage{algorithm}
\usepackage[noend]{algpseudocode}
\usepackage{todonotes}
\usepackage{comment}

\newcommand{\R}{\ensuremath{\mathbb{R}}}

\newcommand{\ra}{\ensuremath{\rightarrow}}

\newcommand{\paren}[1]{\left(#1\right)}

\newcommand{\abs}[1]{\left|#1\right|}

\title{Joint Learning of Domain Classification and Out-of-Domain Detection with Dynamic Class Weighting for Satisficing False Acceptance Rates}
\name{Joo-Kyung Kim and Young-Bum Kim}
\address{
  Amazon Alexa}
\email{\{jookyk, youngbum\}@amazon.com}

\begin{document}

\maketitle
\begin{abstract}
In domain classification for spoken dialog systems, correct detection of out-of-domain (OOD) utterances is crucial because it reduces confusion and unnecessary interaction costs between users and the systems. Previous work usually utilizes OOD detectors that are trained separately from in-domain (IND) classifiers, and confidence thresholding for OOD detection given target evaluation scores. In this paper, we introduce a neural joint learning model for domain classification and OOD detection, where dynamic class weighting is used during the model training to \textit{satisfice} a given OOD false acceptance rate (FAR) while maximizing the domain classification accuracy. Evaluating on two domain classification tasks for the utterances from a large spoken dialogue system, we show that our approach significantly improves the domain classification performance with satisficing given target FARs.
\end{abstract}

\section{Introduction}
\label{sec:intro}
Domain classification is one of the three major components of spoken language understanding along with intent detection and slot filling \cite{Tur2011}. Errors made by domain classifiers are more critical than errors by the other components because the domain classification errors tend to be propagated to completely incorrect system actions or responses. Since recent spoken dialog systems such as Amazon Alexa, Google Assistant, Microsoft Cortana, and Apple Siri deal with a wide variety of scenarios \cite{Sarikaya2016,Sarikaya2017}, correct domain classification of a user’s utterance into one of the supported domains or out-of-domain (OOD) is becoming more complex and important.

Domain classifiers are usually trained focusing on maximizing a single evaluation metric such as classification accuracy and F-score. In real spoken dialog systems, however, correctly detecting OOD utterances\footnote{The OOD explicitly acknowledges systems' inability to respond to the user’s request whenever it does not have a valid response.} is crucial because spoken dialog systems are prone to receive various OOD utterances such as unactionable utterances, ungrammatical utterances, and those with severe ASR errors. Therefore, misclassifying OOD utterances as in-domain (IND) causes confusion and unnecessary interaction costs between users and the dialog systems. To reduce the OOD misclassification of domain classifiers, false acceptance rate (FAR), which is 1 - OOD recall, is often regarded as a \textit{satisficing} metric that must be below a predefined value while false rejection rate (FRR), which is 1 - IND recall, is considered relatively less important.

Previous approaches for OOD detection usually train OOD detectors separately or on top of IND classifiers \cite{Lane2007, Tur2014, Ryu2017, Oh2018}.
In these methods, OOD detectors are trained only to identify whether the given utterances are OOD or not regardless the classification of IND utterances.
Also, the methods are evaluated either based on IND recall and OOD recall \cite{Tur2014} or Equal Error Rate (ERR), where thresholding is used to match FAR and FRR to be the same \cite{Lane2007, Ryu2017}. Consequently, they do not specifically focus on keeping FARs to be low.

Dealing with those issues, we introduce a joint learning model of IND classification and OOD detection. Joint learning models have been shown effective for various spoken language understanding tasks. For example, joint training of intent detection and slot-filling \cite{JKKim2016}, joint training of all the three SLU components \cite{Hakkani-Tur2016, YBKim2017}, and joint training of multiple domains \cite{kim2015new, Jaech2016, kim2016frustratingly} have been shown synergistic since the jointly trained components are highly related to each other. Our model jointly trains a multi-class classifier for the domain classification and a binary classifier for the OOD detection on top of a bidirectional Long Short-Term Memory (BiLSTM) layer \cite{Graves2005} for the utterance representations.
Within this joint architecture, IND classification and OOD detection are helpful to each other by sharing underlying vector representations.
In addition, we use dynamic class weighting, where we adjust the class weights for the IND and OOD loss functions to satisfice the FAR on the development set for each epoch. With dynamic class weighting, we first focus on the FAR as a satisficing metric that must be equal or lower than a predefined target value and then the IND accuracy as an optimizing metric.

Evaluating on two datasets collected by Amazon Alexa, we show that our joint learning model, which aims to satisfice the FAR and maximize the overall classification accuracy with dynamic class weighting, significantly improves domain classification performance given the two metrics.






\section{Satisficing false acceptance rates}
Our objective in this paper is having FAR as a satisficing metric and the domain classification accuracy as an optimizing metric.
This is relevant to addressing class imbalance or unequal class cost cases in classification, where techniques such as oversampling, undersampling, SMOTE, class weighting, and threshold-moving are commonly used \cite{Chawla2002,Japkowicz2002,Maloof2003,Zhou2006,Buda2017}. However, our objective is different from class imbalance or unequal class costs because of the following reasons:
\begin{itemize}
\item{We have two metrics (FAR and accuracy) to optimize rather than one.}
\item{FAR must be satisficed to be equal or below a predefined target value.}
\item{Proper oversampling rates or class weights for satisficing a given FAR is difficult to be decided in advance since they would be substantially different for different datasets.}
\end{itemize}

\begin{figure*}[t]
\centering
\includegraphics[scale=0.43]{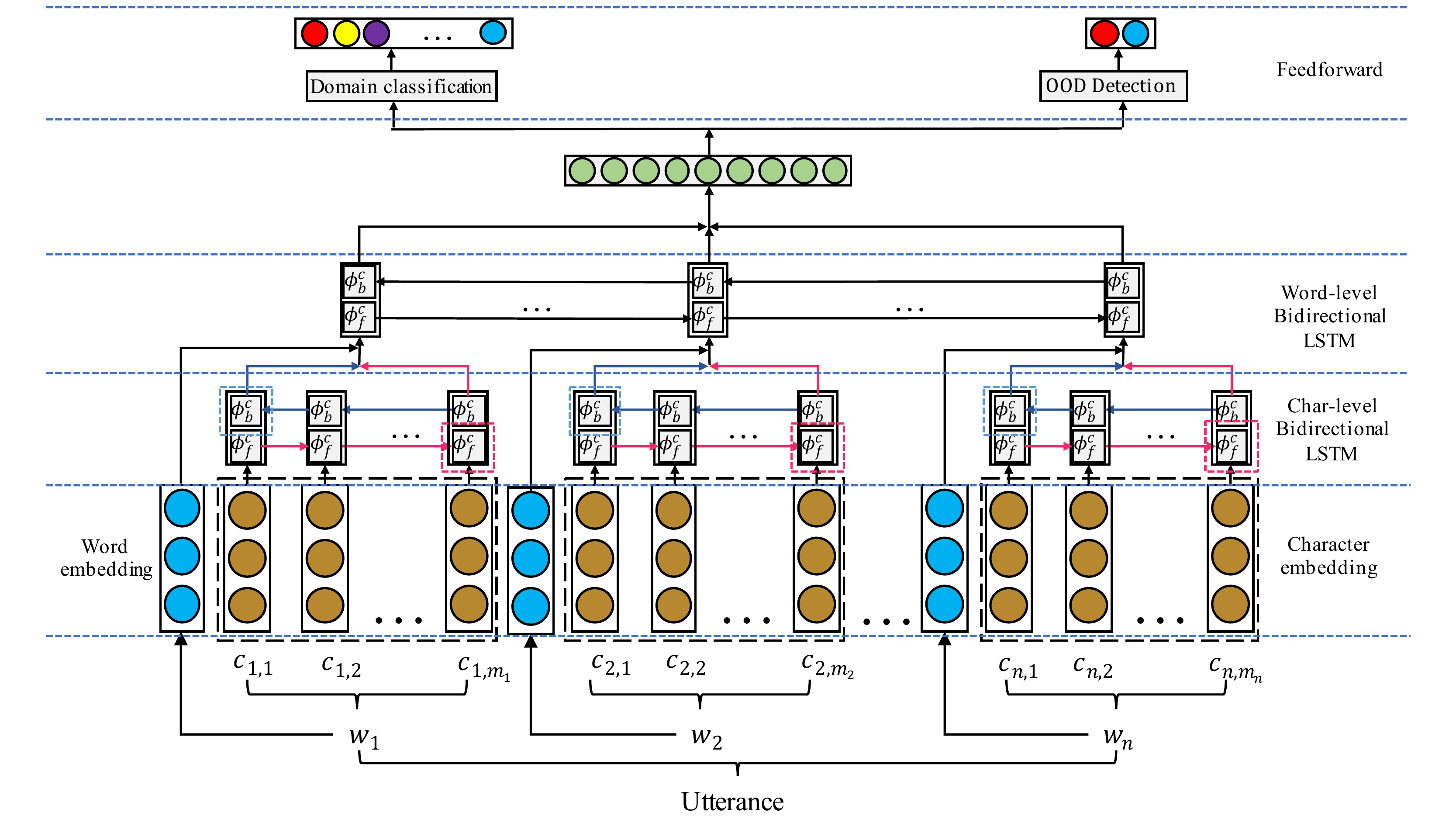}
\caption{\small Model architecture: each word is represented by the concatenation of the word vector from word embedding and the orthography-sensitive vector from the last outputs of two character LSTMs. On top of the word representations, we use a BiLSTM to represent the word sequence as a vector sequence. The last outputs of the two LSTMs are concatenated to be used as a single vector representing the entire utterance. On top of that, we jointly train a domain classifier and an OOD detector.}
\label{fig:joint_domain_detector}
\end{figure*}

We formulate the objective as a non-differentiable constrained optimization problem as follows:
\begin{align}
\max accuracy\left(\mathcal{D}\right) \quad \text{subject to} \quad & FAR\left(\mathcal{D}\right) \leq T,
\label{eq:sfar}
\end{align}
where $\mathcal{D}$ is an evaluated dataset and $T$ is a target FAR.
In Section \ref{sec:model}, we describe our model and formulate a differentiable surrogate loss function to address the objective.

\section{Model}
\label{sec:model}
Figure \ref{fig:joint_domain_detector} shows the overall architecture of the proposed joint model of domain classification and OOD detection.
\subsection{Word representation}
\label{sec:embedding_layer}
In order to leverage character-level vector representations, we use both character embeddings and word embeddings to represent each word \cite{Plank2016}. Let $\mathcal{C}$ and $\mathcal{W}$ denote the set of characters and the set of words, respectively.
Let $\oplus$ denote the vector concatenation operator.
We formulate LSTM \cite{Hochreiter1997} as a function $\phi:\R^d\times\R^{d'} \ra \R^{d'}$
that takes an input vector $x$ and a state vector $h$ to output a new state vector $h' = \phi(x, h)$.

The model parameters associated with the word representations are:
\begin{align*}
\textbf{Char embedding: } &e_c \in \R^{25} \text{ for each } c \in \mathcal{C} \\
\textbf{Char LSTMs: } &\phi^{\mathcal{C}}_f, \phi^{\mathcal{C}}_b: \R^{25} \times \R^{25} \ra \R^{25} \\
\textbf{Word embedding: } &e_w \in \R^{100} \text{ for each } w \in \mathcal{W}
\end{align*}

Let $\left(w_1, \ldots, w_n\right)$ denote a word sequence where word $w_i$ has a character $w_i(j) \in \mathcal{C}$ at position $j$.
The vector representation of the $i$-th word, $v_i \in \R^{150}$, is obtained by the concatenation of the both ends of the character LSTM outputs and the $i$-th word vector $e_{w_i}$.\footnote{$e_{w}$ is pretrained with GloVe leveraging Wikipedia 2014 and Gigaword 5 corpora \cite{Pennington2014}.} It is formulated as follows:
\begin{align*}
f^{\mathcal{C}}_j &= \phi^{\mathcal{C}}_f \paren{e_{w_i\left(j\right)}, f^{\mathcal{C}}_{j-1}} &&\forall j = 1 \ldots \abs{w_i} \\
b^{\mathcal{C}}_j &= \phi^{\mathcal{C}}_b\paren{e_{w_i\left(j\right)}, b^{\mathcal{C}}_{j+1}} &&\forall j = \abs{w_i} \ldots 1\\
v_i &= f^{\mathcal{C}}_{\abs{w_i}} \oplus b^{\mathcal{C}}_1 \oplus e_{w_i} &&,
\end{align*}
where $f^{\mathcal{C}}_{\abs{w_i}}$, $b^{\mathcal{C}}_1$ denote the forward LSTM output for the last character and the backward LSTM output for the first character, respectively.

\subsection{Utterance representation}
\label{sec:uttrep}
We encode the word vector sequence $\left(v_1,\ldots,v_n\right)$ with a BiLSTM for each $i = 1,\ldots,n$.\footnote{$f^{\mathcal{C}}_0, b^{\mathcal{C}}_{\abs{w_i}+1}, f^{\mathcal{W}}_0$, and $b^{\mathcal{W}}_{\abs{n}+1}$ are randomly initialized vectors.}:
\begin{align*}
\textbf{Word LSTMs: } &\phi^{\mathcal{W}}_f, \phi^{\mathcal{W}}_b: \R^{150} \times \R^{100} \ra \R^{100}.\\ 
\end{align*}
\vspace{-7mm}
\begin{align*}
f^{\mathcal{W}}_i &= \phi^{\mathcal{W}}_f\paren{v_i, f^{\mathcal{W}}_{i-1}}  &&\forall i = 1 \ldots n \\
b^{\mathcal{W}}_i &= \phi^{\mathcal{W}}_b\paren{v_i, b^{\mathcal{W}}_{i+1}} &&\forall i = n \ldots 1.
\end{align*}

Then, in a similar way to obtaining the word representations from the character BiLSTMs, we concatenate the last outputs of both word LSTMs to represent the whole utterance as a single 200 dimensional vector:
\begin{align*}
u &= f_n^\mathcal{W} \oplus b_1^\mathcal{W}.
\end{align*}

We also evaluate two other utterance representation methods, which are word vector summation $u=\sum_{i=1}^{n}v_i$ and convolutional neural networks (CNN).\footnote{In a similar method to \cite{Kim2014}, we use three convolution filters whose sizes are 3, 4, and 5 on top of a word vector sequence. Then, we use max pooling for each filter output, and finally concatenate the three max pooling outputs to represent the whole utterance.}

\subsection{Domain classification}
On top of the utterance vector $u$, we use a feed-forward neural network $\phi_{d}$ and softmax function to obtain the probability distribution over the entire domains\footnote{We use a single hidden layer with SeLU activation function \cite{Klambauer2017} for normalized activation outputs.} as:
\begin{align*}
d &= softmax\left(\phi_d\left(u\right)\right).
\end{align*}
The loss function for the domain prediction is formulated as the cross-entropy between the label and the probability distribution of an utterance:
\begin{equation}
\mathcal{L}_D\left(d\right) = - \hat{d}\log d,
\end{equation}
where $\hat{d}$ is the one-hot representation of the ground-truth domain of the current utterance.


\subsection{OOD detection}
Along with the domain classifier, we jointly train an OOD detector, which predicts whether the current utterance is IND or OOD. We use a feed-forward network $\phi_{o}$ as did for the domain classification:
\begin{align*}
o &= softmax\left(\phi_{o}\left(u\right)\right).
\end{align*}
The loss function for OOD detection is formulated as:
\begin{equation}
\mathcal{L}_O\left(o\right) = - \hat{o}\log o,
\end{equation}
where $\hat{o}$ is a two dimensional one-hot vector representing whether the current utterance is IND or OOD.

\subsection{Joint loss function}
\label{sec:jointlossfunc}
A loss function for combining domain classification and OOD detection is formulated as follows:
\begin{equation}
\mathcal{L}_J\left(\cdot\right) = \mathcal{L}_D\left(\cdot\right) + \alpha\mathcal{L}_O\left(\cdot\right),
\label{eq:jointloss}
\end{equation}
where 
$\alpha$ is a hyperparameter that controls the degree of the influence from the binary OOD detector.
We show results on different $\alpha$ values in Section \ref{ssec:results}.

\subsection{Dynamic class weighting}
\label{sec:dynamic_class_weighting}
The final loss function, which approximately optimizes Equation \ref{eq:sfar}, is as follows:
\begin{equation}
\mathcal{L}\left(\mathcal{D}\right) = \left(2-\lambda\right)\sum_{d_k \in \mathcal{D}_{IND}} \mathcal{L}_J\left(d_k\right)  + \lambda \sum_{d_l \in \mathcal{D}_{OOD}} \mathcal{L}_J\left(d_l\right),
\label{eq:dynamicweight}
\end{equation}
where $\mathcal{D}_{IND}$ is the set of utterances with IND ground-truths, $\mathcal{D}_{OOD}$ is the utterance set with OOD ground-truths, $\mathcal{L}_J$ is the joint loss function in Section \ref{sec:jointlossfunc}, and $\lambda$ is a parameter deciding the class weights for IND domains and the OOD.
Here, $\sum_{d_k \in \mathcal{D}_{IND}} \mathcal{L}_J\left(d_k\right)$ and $\sum_{d_l \in \mathcal{D}_{OOD}} \mathcal{L}_J\left(d_l\right)$ are surrogate loss functions for maximizing the IND classification and satisficing the FAR, respectively.
This formulation uses $2-\lambda$ and $\lambda$ as the class weights for IND and OOD, respectively. The main issue of Equation \ref{eq:dynamicweight} is that we cannot predetermine $\lambda$ as aforementioned.

To obtain a proper $\lambda$, we introduce dynamic class weighting for OOD, where $\lambda$ is changed during the training so that the FAR on the development set is satisficed with minimal $\lambda$ increase for OOD.
We initialize $\lambda$ to 1 so that the class weights for both IND and OOD are 1 in the beginning. At the end of each training epoch, we calculate the FAR on the development set. If the current FAR does not satisfice the target FAR, we add $\gamma$ to current $\lambda$. Oppositely, if the target FAR is satisficed, we subtract $\gamma$ from current $\lambda$. When adding and subtracting $\gamma$, we limit the $\lambda$ value to be less than 2 and larger than 0. In our work, we initialize $\gamma$ to 0.1.\footnote{We also tried different values but there were no significant differences in the experiment results.} To reduce fluctuations of $\lambda$ during the late epochs, we halve $\gamma$ each time when the target FAR is satisficed in the current epoch but not in the previous epoch.
With this approach, we encourage the model to find the minimal class weight change that satisfices the FAR and then focuses on the overall classification accuracy.

\section{Experiments}
\label{sec:exp}
We have conducted a series of experiments to evaluate the proposed method on datasets obtained from real usage data in Amazon Alexa with two domain classification tasks.

\subsection{Datasets}
We evaluate our models on two domain classification tasks from different data sources: (1) utterances from 21 Alexa domains, (2) utterances from frequently used 1,500 skills out of more than 40,000 skills.\footnote{In Amazon Alexa, a \textit{skill} is a domain developed by third-party developers \cite{Kumar2017, YBKim2018b}.} For both cases, we use randomly sampled unique utterances that are collected and annotated from the real user logs. The average utterance lengths are 5.96 and 5.68 for the 21 domains and the 1,500 skills, respectively.
Table \ref{tab:dataset_statistic} shows the statistic of the datasets.
\begin{table}[t]
\centering
\begin{tabular}{l|rrr|rrr}
      & \multicolumn{3}{c|}{21 domain dataset}                                                & \multicolumn{3}{c}{1,500 skill dataset}                                              \\ \cline{2-7} 
      & \multicolumn{1}{c}{IND} & \multicolumn{1}{c}{OOD} & \multicolumn{1}{c|}{Total} & \multicolumn{1}{c}{IND} & \multicolumn{1}{c}{OOD} & \multicolumn{1}{c}{Total} \\ \hline
Train & 712k                    & 255k                    & 967k                       & 372k                    & 381k                    & 753k                      \\
Dev   & 112k                    & 17k                     & 129k                       & 103k                    & 35k                     & 138k                      \\
Test  & 112k                    & 21k                     & 133k                       & 105k                    & 35k                     & 104k                      \\ \hline
\end{tabular}
\caption{The numbers of the utterances in the two datasets: 21 Alexa domains and 1,500 Alexa skills.}
\label{tab:dataset_statistic}
\vspace{-7mm}
\end{table}

\subsection{Results}
\label{ssec:results}
Each evaluated model is trained for 50 epochs and we use the parameters at the epoch showing the best score on the development set to report the scores on the test set.
We use ADAM \cite{Kingma2015} with learning rate 0.001 for the optimization. For stable training, we use gradient clipping, where the threshold is set to 5. For efficiency, we use a variant of LSTM, where the input gate and the forget gate are coupled and peephole connections are used \cite{Gers2000, Greff2017}. For the LSTM regularization, we use variational dropout \cite{Gal2016}. All the models are implemented with DyNet \cite{Neubig2017}.

Table \ref{tab:domainresults} and \ref{tab:skillresults} show the classification accuracies on the two datasets with various models given different target FARs. Even though the FAR on the development set is satisficed with dynamic class weighting, the FAR on the test set might not be satisficed. 
To satisfice each given target FAR for the test set, we set a decision threshold to regard a predicted domain as OOD when the highest confidence score is below the threshold.

In Table \ref{tab:domainresults} and \ref{tab:skillresults}, \textit{Separate} models use separate underlying utterance representations for IND only classification and OOD detection. In this case, given an utterance, we first run the OOD detector to predict whether the utterance belongs to IND or OOD. If it is predicted as IND, we run the IND classifier to predict the domain of the utterance.\footnote{Since the OOD detector is solely trained in \textit{Separate} models, we do not need to set $\alpha$ to be relatively low.}

\textit{Joint} models share the underlying utterance representations for both domain classification and OOD detection. The domain classifier also includes OOD as one of the domains so that the domain classifier can also learn representations from OOD utterances.
As aforementioned in Section \ref{sec:uttrep}, we also evaluate the other utterance representation methods, word vector summation and CNN.
We utilize the OOD detector with setting $\alpha$, which is the coefficient for the OOD detection loss in Equation \ref{eq:jointloss}.

\textit{Joint with dynamic class weighting} models are trained including dynamic class weighting with given target FARs for the development sets.

\subsubsection{21 Alexa Domains}
This task classifies input utterances to either one of 21 Alexa domains or OOD. For example, the domains for ``\textit{What's the weather this weekend in Orlando},'' ``\textit{Get me a ride to Seattle airport},'' and ``\textit{Oh no nothing}'' should be classified as \texttt{Weather},  \texttt{BookingAndReservations}, and \texttt{OOD}, respectively.

When no target FAR is given, the accuracy and the FAR of the \textit{Joint (BiLSTM)} model with $\alpha=0$ for this dataset are 91.06\% and 9.75\%, respectively.

We evaluate the proposed models with 6\%, 5.5\%, and 5\% as the target FARs.
Table \ref{tab:domainresults} shows the model evaluation results. 

\begin{table}
\centering
\begin{tabular}{l|l|lll}
\multirow{2}{*}{Model} & \multirow{2}{*}{$\alpha$} & \multicolumn{3}{c}{Target FAR} \\
 &  & 6\% & 5.5\% & 5\% \\ \hline
Separate (BiLSTM) & 1 & 85.69 & 84.68 & 83.7 \\ \hline
Joint (WordVecSum) & 0 & 87.28 & 86.85 & 86.37 \\
Joint (CNN) & 0 & 88.61 & 88.18 & 87.83 \\ \hline
\multirow{5}{*}{Joint (BiLSTM)} & 0 & 89.2 & 88.73 & 88.27 \\
 & 0.001 & 89.33 & 88.94 & 88.4 \\
 & 0.005 & 89.25 & 88.84 & 88.36 \\
 & 0.01 & 89.27 & 88.89 & 88.38 \\
 & 0.05 & 89.26 & 88.88 & 88.43 \\ \hline
\multirow{5}{*}{\begin{tabular}[c]{@{}l@{}}Joint (BiLSTM)\\ w/ dynamic\\ class weighting\end{tabular}} & 0 & 90.67 & 90.52 & 90.26 \\
 & 0.001 & 90.67 & 90.52 & 90.34 \\
 & 0.005 & \textbf{90.71} & 90.53 & \textbf{90.38} \\
 & 0.01 & 90.69 & \textbf{90.61} & 90.28 \\
 & 0.05 & 90.63 & 90.53 & 90.30 \\ \hline
\end{tabular}
\caption{The test classification accuracies (\%) of various models given different satisficing FARs on the 21 domain dataset. $\alpha$ of Equation \ref{eq:jointloss} is set to 0 for \textit{Separate} case and 1 for all the other cases. $\alpha$ is the coefficient for the OOD detector loss.}
\label{tab:domainresults}
\vspace{-7mm}
\end{table}

Since domain classification and OOD detection are closely related tasks, it is shown that \textit{Joint (BiLSTM)} models outperform \textit{Separate (BiLSTM)} model for all the cases. Also, to represent the utterances in a vector space, using BiLSTM is shown to be consistently better than using word vector summation and CNN in our experiments.

For \textit{Joint} models, utilizing the OOD detector by setting $\alpha$ to be higher than 0 during the training shows better accuracies than not using it. This demonstrates that jointly training a separate OOD detector noticeably helps increase the overall classification performance.

We can observe that the accuracies of \textit{Joint with dynamic class weighting} models are significantly higher than those of the other models. This shows that utilizing dynamic class weighting is effective for our objective, where we first satisfice a given FAR and then maximize the accuracy by dynamically finding more effective class weights.

\subsubsection{1,500 Alexa Skills}
This task deals with utterance classification to either one of 1,500 skills or OOD. For example, the skills for ``\textit{what does a peacock say}'' and ``\textit{find me the recipe for world’s best lasagna}'' should be predicted as \texttt{ZooKeeper} and \texttt{AllRecipes}, respectively. The skills are significantly more diverse and less well defined than 21 Alexa domains, which makes the classification more challenging. 
In real spoken dialog systems, the classification performance can be further improved by leveraging various contextual information \cite{Robichaud2014, Crook2015, YBKim2018a, YBKim2018b}. However, they are beyond the scope of this paper, and we leave the evaluation of our models on such reranking systems as future work.

On this task, when there is no target FAR, the accuracy and the FAR of the \textit{Joint (BiLSTM)} model with $\alpha=0$ are 80.65\% and 3.62\%, respectively.

Therefore, we have evaluated our models on lower target FARs, 2\%, 1.5\%, and 1\%. Table \ref{tab:skillresults} shows the results of our proposed models.
\begin{table}
\centering
\begin{tabular}{l|l|lll}
\multirow{2}{*}{Model} & \multirow{2}{*}{$\alpha$} & \multicolumn{3}{c}{Target FAR} \\
 &  & 2\% & 1.5\% & 1\% \\ \hline
Separate (BiLSTM) & 1 & 77.50 & 75.40 & 72.41 \\ \hline
Joint (WordVecSum) & 0 & 74.07 & 72.31 & 69.28 \\
Joint (CNN) & 0 & 77.69 & 75.95 & 73.50 \\ \hline
\multirow{5}{*}{Joint (BiLSTM)} & 0 & 78.19 & 76.48 & 74.10 \\
 & 0.001 & 78.37 & 76.97 & 74.32 \\
 & 0.005 & 78.14 & 76.59 & 74.25 \\
 & 0.01 & 78.32 & 76.44 & 74.34 \\
 & 0.05 & 78.05 & 76.36 & 73,97 \\ \hline
\multirow{5}{*}{\begin{tabular}[c]{@{}l@{}}Joint (BiLSTM)\\ w/ dynamic\\ class weighting\end{tabular}} & 0 & 79.18 & 78.26 & 76.74 \\
 & 0.001 & 79.26 & 78.63 & 77.22 \\
 & 0.005 & \textbf{79.34} & \textbf{78.91} & \textbf{77.32} \\
 & 0.01 & 79.20 & 78.54 & 77.05 \\
 & 0.05 & 79.13 & 78.27 & 77.07 \\ \hline
\end{tabular}
\caption{The test classification accuracies (\%) on the 1,500 skill dataset.}
\label{tab:skillresults}
\vspace{-7mm}
\end{table}
Overall, similarly to the results of 21 Alexa Domains, we can see that \textit{Joint} models are better than \textit{Separate} model, using BiLSTM outperforms using word vector summation or CNN for the utterance representations, and \textit{Joint with dynamic class weighting} models show significantly better performance than other models.

\section{Conclusion}
We have introduced a joint learning model of domain classification and OOD detection utilizing dynamic class weighting to satisfice a target FAR and then maximize the overall classification accuracy. Evaluating on two domain classification tasks for the utterances from Amazon Alexa, we have shown that our proposed joint learning models with dynamic class weighting is more effective than the models with separate learning of domain classification and OOD detection or those trained to optimize a single metric when we have FAR as a satisficing metric and accuracy as an optimizing metric.

\bibliographystyle{IEEEtran}

\bibliography{refs}


\end{document}